# PEFT-MedSAM: Efficient Fine-Tuning of Medical Foundation Models for Explainable Skin Lesion Segmentation

**Asad Channa [1], Abdullah Khan [1], Asghar Ali Chandio[2], Aamir Akbar [3], Shahzad Memon[4,*], Aqib Hussain [1], Ameer Hamza[1]**

[1]Department of Computer Science, Quaid-e-Awam University of Engineering, Sciences & Technology, Email: drasadchanna657@gmail.com

[2]Department of Artificial Intelligence, Quaid-e-Awam University of Engineering, Sciences & Technology, Pakistan, Email: asghar.ali@quest.edu.pk

[3]Department of Computer Science, Sindh Madressatul Islam University, City Campus, Karachi

[4]Department of Computer Science and Digital Technologies, School of Architecture, Computing and Engineering, University of East London, Email: S.Memon@uel.ac.uk

***** Correspondence: S.Memon@uel.ac.uk

**Abstract:** Automated segmentation of skin lesions using deep learning models for dermoscopic images can be very helpful in finding melanomas earlier than they would normally be detected. However, most deep learning methods available do not perform well. The aim of this paper is to present a parameter-efficient fine-tuning method called PEFT-MedSAM for adapting the Medical Segment Anything Model (MedSAM) to automatically segment dermoscopic skin lesions. The PEFT-MedSAM method uses only the lightweight mask decoder for training the model while keeping the pre-trained image encoder and prompt encoder frozen. The experiments performed on the ISIC 2018 benchmark dataset shows that PEFT-MedSAM obtains a dice coefficient of .9411 and an intersection over union value of .8918 when compared to both a fully trained U-Net baseline (.8715 dice coefficient) and zero-shot MedSAM inference (.8997 dice coefficient). The external validation of the model using PH2 dataset shows .9467 dice coefficient with +/- .0310 standard deviation. Supportive evidence for these claims include a p-value less than .0001 for Wilcoxon signed rank tests comparing the two datasets and bootstrap-estimated 95% confidence intervals of [.9364,.9447] that represent the estimated range of possible values for the average dice coefficient obtained by repeating the test. To increase clinical trustworthiness, we used Grad-CAM explainability along with a pointing game based evaluation methodology to evaluate the CNN baseline model on the validation set. The results showed that we had an accuracy rate of 98.27% on the validation set of 519 images and confirmed that the model classified regions containing skin lesions.

---

## 1. Introduction

Skin cancer is one of the most common cancers in the world, with an estimated of 5.4 million new cases being diagnosed in the United States of America (USA) annually [1, 2] and the global burden is expected to rise to 510,000 by 2040 [3]. Although melanoma is only a small percentage of the total number of skin cancer cases, it is responsible for the majority of skin cancer deaths [21]. The prognosis for melanoma is very strongly stage dependent: five-year survival rates are >99% for early, localized disease, but drop sharply to ~25% for distant metastasis [4]. Dermoscopy has become the clinical gold standard for non-invasive examination of skin lesions, allowing the visualization of sub surface morphological structures by polarized or immersion light microscopy [5]. To accurately delineate lesion boundaries from dermoscopic images is essential for quantitative analysis of diagnostic features such as asymmetry, border irregularity, color variation, and diameter (ABCD rule) which are the accepted criteria for the assessment of melanoma [6, 22]. Furthermore, manual delineation by dermatologists is also time-consuming, subjective and prone to inter and intra observer variability [20]. Disagreements among experienced dermatologists regarding unclear boundaries of lesions have been reported up to 30% [7] and automated, objective computational methods are needed to complement clinical decision making at scale.

The International Skin Imaging Collaboration (ISIC) challenges have set a high benchmark for skin lesion segmentation using deep convolutional neural networks (CNNs), especially those based on the U-Net architecture [8]. But the architectures trained from scratch for tasks have three fundamental drawbacks. First, they need a large amount of annotated training data, which is difficult to be acquired in clinical settings because of the need for highly trained individuals to accurately delineate lesions [9]. Second, they do not exploit the wealth of visual representations that are already present in the large-scale pre-trained models, and they are restricted to data distributions not observed at training time [10]. Third, the existing literature has primarily focused on segmentation accuracy, neglecting the need for clinical interpretability, or the capacity to explain the model's segmentation decision, which is becoming a key prerequisite for regulatory clearance and a clinician's trust of an AI-powered diagnostic system [11, 12, 13].

Recently, the emergence of vision foundation models, such as the Segment Anything Model (SAM) [14] and medical adaptation MedSAM, has marked a paradigm shift to universal segmentation systems trained on datasets with unprecedented scale. MedSAM was trained with more than 1.5 million medical image-mask pairs from across 10 medical imaging modalities [15] and showed promising zero-shot generalization across a variety of medical segmentation tasks. Yet, despite their impressive capabilities, foundation models have a number of challenges in specialized clinical deployment. Most research institutions are unable to perform full fine-tuning of MedSAM's 93M parameters, and catastrophic forgetting may occur if the learned representations are forgotten [16]. Moreover, the interpretability of the prediction of the foundation model in clinical scenarios is not well discussed in the literature [12].

To overcome these essential limitations, in this paper, we introduce a parameter-efficient fine-tuning framework tailored for dermoscopic skin lesion segmentation, PEFT-MedSAM. The motivation for our approach is that the image encoder pre-trained on a variety of medical imaging data already encodes rich and transferable visual representations, which we hope to leverage for MedSAM. PEFT-MedSAM is task-specific adapted by freezing the image encoder and prompt encoder, and only fine-tuning the lightweight mask decoder, which significantly reduces the computation resources while retaining the general knowledge of the pre-trained image encoder [17].

In addition to segmentation performance, this paper examines the attention behaviour of the CNN baseline with the help of the Pointing Game metric [19] and an additional method: Gradient-weighted Class Activation Mapping (Grad-CAM) [18]. This analysis shows objective and quantitative evidence that attention maps correspond to clinically relevant lesion regions, thereby providing complementary interpretability insights and by directly addressing the trustworthiness requirements that are common and growing expectations in clinical deployment of AI systems.

The major contributions of this paper are as follows:

1. We propose PEFT-MedSAM, a novel parameter-efficient fine-tuning (PEFT) strategy for MedSAM which fine-tunes only 4.3% of the model parameters and yet achieves a Dice score of 0.9411 on ISIC 2018, which is 7.99% higher than the Dice score of the fully trained U-Net baseline and 4.60% higher than zero-shotting MedSAM inference.

2. We systematically compared task-specific CNN (U-Net), zero-shot foundation model (MedSAM), and parameter-efficiently fine-tuned foundation model (PEFT-MedSAM) in dermoscopic skin lesion segmentation and statistically validated evidence (Wilcoxon $p < 0.0001$) of their performance superiority.

3. We present quantitative explainability evaluation using the Grad-CAM analysis and the Pointing Game metric over the CNN baseline, with 98.27% accuracy on 519 validation images and with a promising result of lesion-focused attention localisation. The limitations of gradient-based saliency approaches in Vision Transformer models are discussed, and the directions of future research are identified.

4. We validate cross-dataset generalizability through external evaluation on the independent PH2 dermoscopic dataset, achieving Dice of $0.9467 \pm 0.0310$ on 200 completely unseen images from a different imaging protocol and patient population.

## 2. Related Work

### *2.1. Skin Lesion Segmentation*

Automated segmentation of skin lesions from dermoscopic images has been extensively investigated using deep learning, with the encoder-decoder U-Net architecture [8] establishing the foundational framework through skip connections that preserve spatial detail essential for precise boundary delineation. The next step has been to build upon this foundation with attention mechanisms, hybrid CNN-transformer architectures and ensemble approaches to overcome the intrinsic difficulties of lesion shape variability, low-contrast boundaries, and hair occlusion.

Attention-based extensions have shown to be consistently better than the standard U-Net. Ahmadi et al. [23] proposed a Multiscale Attention U-Net, which uses multi-scale attention at the bottleneck and a Bidirectional Convolutional LSTM, achieving a DSC of 0.937 on PH2. The W-shaped dual encoder-decoder was proposed by Khouloud et al. [24] which was trained together with an Inception-ResNet classifier, achieving a Dice of 93% on the ISIC 2018. Both methods give competitive results, but neither offers explainability mechanisms nor uses pre-trained representations. To learn local texture and global context, hybrid CNN-transformer architectures have been developed. Karimi et al. [25] proposed DEU-Net, which consists of parallel CNN and transformer encoders, and obtained Dice of 90.81% on ISIC 2018 and 95.65% on PH2. Tasnim and Ahamed [26] added Squeeze-and-Excitation blocks and Atrous Spatial Pyramid Pooling to UNet++. They assessed their model just with ISIC 2017. These task-specific architectures, although performing well, demand huge amounts of annotated data and lack interpretability for clinical use.

Efficiency-focused and ensemble approaches are complementary. Hasan et al. [27] proposed DSNet that uses depth-wise separable convolutions to reduce the number of parameters, achieving a mean IoU of 87.0% on PH2 without an evaluation on ISIC 2018. Goyal et al. [28] fused Mask R-CNN and DeepLabV3+ to obtain an ensemble model with a high of 89.93% sensitivity on ISIC 2017, but with significant computational complexity. The FPN+U-Net combination is only evaluated on the ISIC 2016 dataset which is limited to this benchmark only, and has been compared to the Dice 0.93 by Sharen et al. [29]. Although these advances have been achieved, there are three related points of missing research in the current literature: task-specific training without pre-trained representations, inconsistent cross-dataset validation, and lack of quantitative evaluation of explainability. This present work tackles all three limitations head-on.

### *2.2. Foundation Models in Medical Imaging*

With the advent of the foundation models, specifically the Segment Anything Model (SAM), medical image segmentation without task-specific training has taken new directions. Ma et al. [15] introduced MedSAM, a fine-tuned SAM trained on 1,570,000 image–mask pairs from 10 medical imaging modalities and evaluated with 86 in-house and 60 out-of-sample medical image–mask pairs, beating modality specific specialists on the 60 out-of-sample tasks. Ali et al. [30] and Berrezueta et al. [31] report that SAM has potential, but it is not for medical data and remains difficult to deploy in the clinic. For that, M. Wei et al. [32] proposed Rep MedSAM, which replaces the heavy image encoder with lightweight RepViT and apply knowledge distillation, successfully obtained 85.9% average DSC, 2.41 s per case on CPU. Another work, I-MedSAM by Wei et al. [33] adds an implicit neural representation decoder and also a high frequency adapter, with just 1.6 M parameters learnt to enhance boundary detail and cross domain skills. In spite of the improvements, there is no current foundation model adaptation tailored for dermoscopic skin lesion segmentation specifically with quantitative explainability, which is covered by the proposed PEFT MedSAM framework. We note that other foundation models, such as SEEM have performed well on general segmentation benchmarks [34-35], for which they have not been applied to dermatological images. With regard to the vision transformer, we seek to follow the same principle, applying parameter efficient transfer learning techniques, such as adapters or reparameterization, for adapting a large model with minimal computational cost, which is what we do in PEFT MedSAM. A review of the deep learning based image segmentation techniques for the bone fracture in X-Ray images is given in [36].

### *2.3. Explainable AI in Dermatology*

Interpretability is gaining importance in the context of clinical use of AI dermatological tools. In a few studies, Explanable AI (XAI) has been applied to classify skin lesions. Nigar et al. [37] implemented ResNet 18 with visual explanations for ISIC 2019 and achieved 94.47% accuracy. Aras and Doğan [38] have compared the performance of XceptionNet, DenseNet and MobileNet on HAM10000 and found that XAI improves clinical trust when used in conjunction with Grad CAM and LIME. Alrabai et al. [39] assessed four XAI models: InceptionV3, Xception, ResNet50V2, and DenseNet121, which yielded an accuracy of 90.15% using the Xception model, and they found that Xception builds clinician trust. In another study by Mahmud et al. [40] a custom CNN was used with Grad CAM and saliency maps and resulted in over 95% accuracy on two datasets, and another study with 76 dermatologists showed that heatmaps helped dermatologists make more accurate diagnoses. Grad CAM, LIME and occlusion sensitivity were applied to ResNet 50, GoogLeNet and SqueezeNet by Khan et al. [41] and they noted that all the XAI techniques indicated clinically relevant lesion centers, color differences and irregular borders. Finally, Munjal et al. (SkinSage XAI) [42] used Inception V3 and Grad CAM and LIME on 50,000 images of HAM10000, which yielded 96% accuracy and AUC of 99.83%. It is remarkable that all the above mentioned works focus on classification; however, segmentation, and in particular the quantitative explainability for the segmentation of dermoscopic lesion boundaries, is under-explored. The present study addresses this gap by integrating Grad-CAM with the Pointing Game metric on the CNN baseline to provide objective, measurable evidence of attention alignment with ground truth lesion regions, while also discussing the implications of applying gradient-based saliency methods to transformer-based architectures.

## 3. METHODOLOGY

### *3.1. Overview*

The overall proposed pipeline is illustrated in Figure. 1. The input dermoscopic images are first preprocessed and then processed by three different models: U-Net trained from scratch, MedSAM in zero-shot configuration, and PEFT-MedSAM with parameter-efficient fine-tuning. Dice coefficient, Intersection over Union (IoU) and explainability scores are used for all models on two separate dermoscopic datasets.

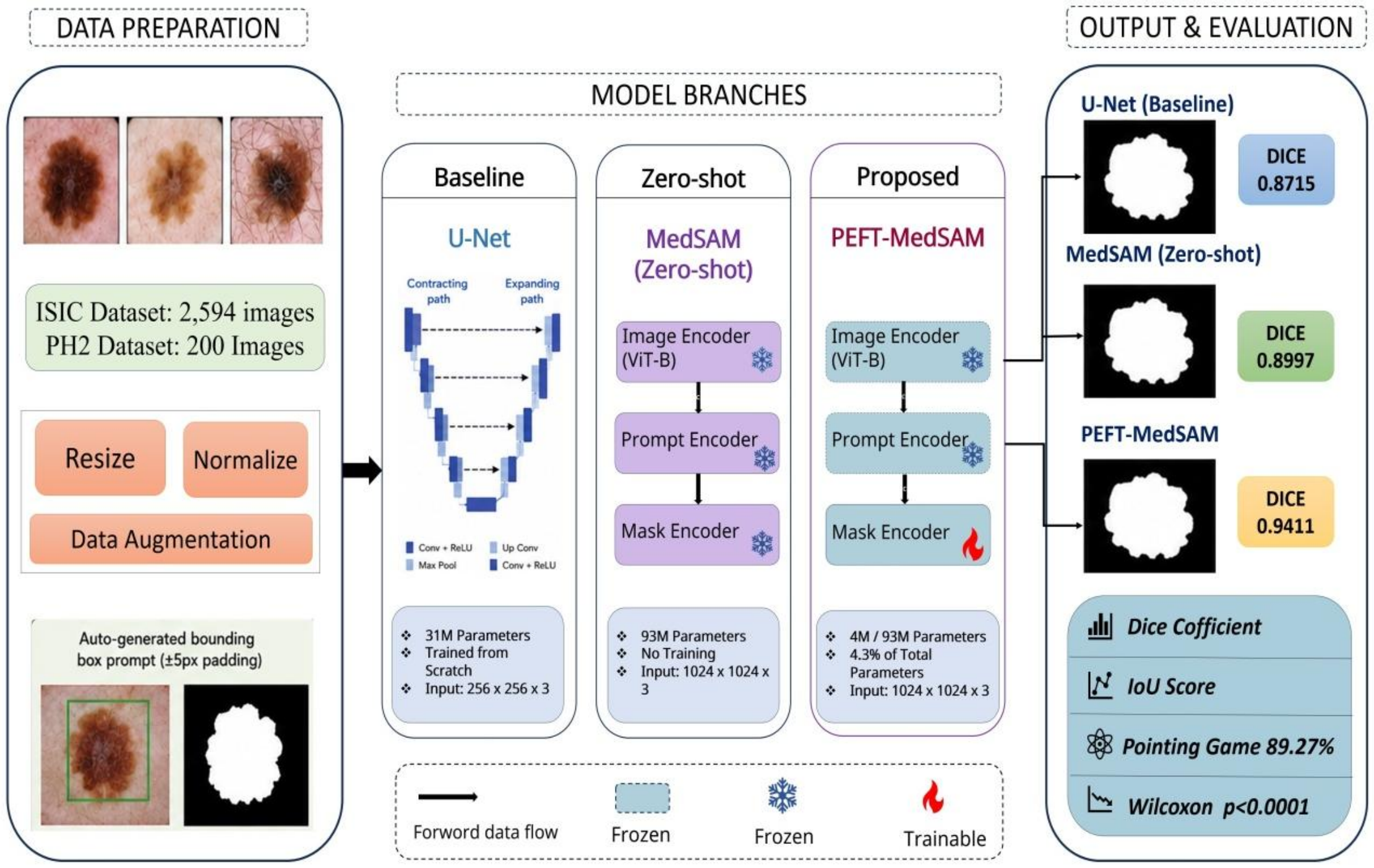

**Figure 1. PEFT-MedSAM pipeline.** Data preparation → three parallel models (U-Net, MedSAM zero-shot, PEFT-MedSAM) → segmentation outputs → evaluation metrics (Dice, IoU, Pointing Game, Wilcoxon). Best performance is achieved by the proposed PEFT-MedSAM (Dice 0.9411, $p < 0.0001$).

### ***3.2. Datasets***

#### 3.2.1 ISIC 2018

The main dataset used for this study is the International Skin Imaging Collaboration (ISIC) 2018 [43] Task 1. A total of 2,594 dermoscopic images were captured with varying clinical settings and a binary segmentation mask created showing the boundaries of the lesions manually annotated by board certified dermatologists. Lesions show a high degree of size, shape and colour variation as well as varying backgrounds that represent the wide variety of clinical presentations, with some exhibiting very light background skin colour and others darker. The data was then split as usual, with 2,075 samples for training and 519 for validation, and with a fixed random seed to ensure the results are reproducible.

### 3.2.2 PH2 Dataset

To evaluate the generalizability of the model to completely new data, only the external validation set (PH2) was used [44]. PH2 contains 200 dermoscopic images of melanocytic lesions with 80 common nevi, 80 atypical nevi and 40 melanomas, with expert annotations of every image showing the segmentation. The unique protocol, imaging data, and patient cohort used in the acquisition of the PH2 data set represents a stringent challenge to the cross-dataset generalizability of ISIC 2018. No PH2 images were utilized in the training of the model and in hyperparameter selection. The statistics of the datasets are summarized in Table 1.

**Table 1.** Summary of dermoscopic datasets used for model training and evaluation.

| Dataset | Total Images | Resolution | Lesion Types | Source | Training | Validation |
|---|---|---|---|---|---|---|
| ISIC 2018 | 2,594 | Variable | Melanocytic lesions | ISIC Archive | 2075 | 519 |
| PH2 | 200 | 768 × 560 px | Common nevi, Atypical nevi, Melanoma | Hospital Pedro Hispano, Portugal | - | - |

Figure 2 illustrates a representative set of samples from the two datasets with original dermoscopic images and their corresponding ground truth segmentation masks.

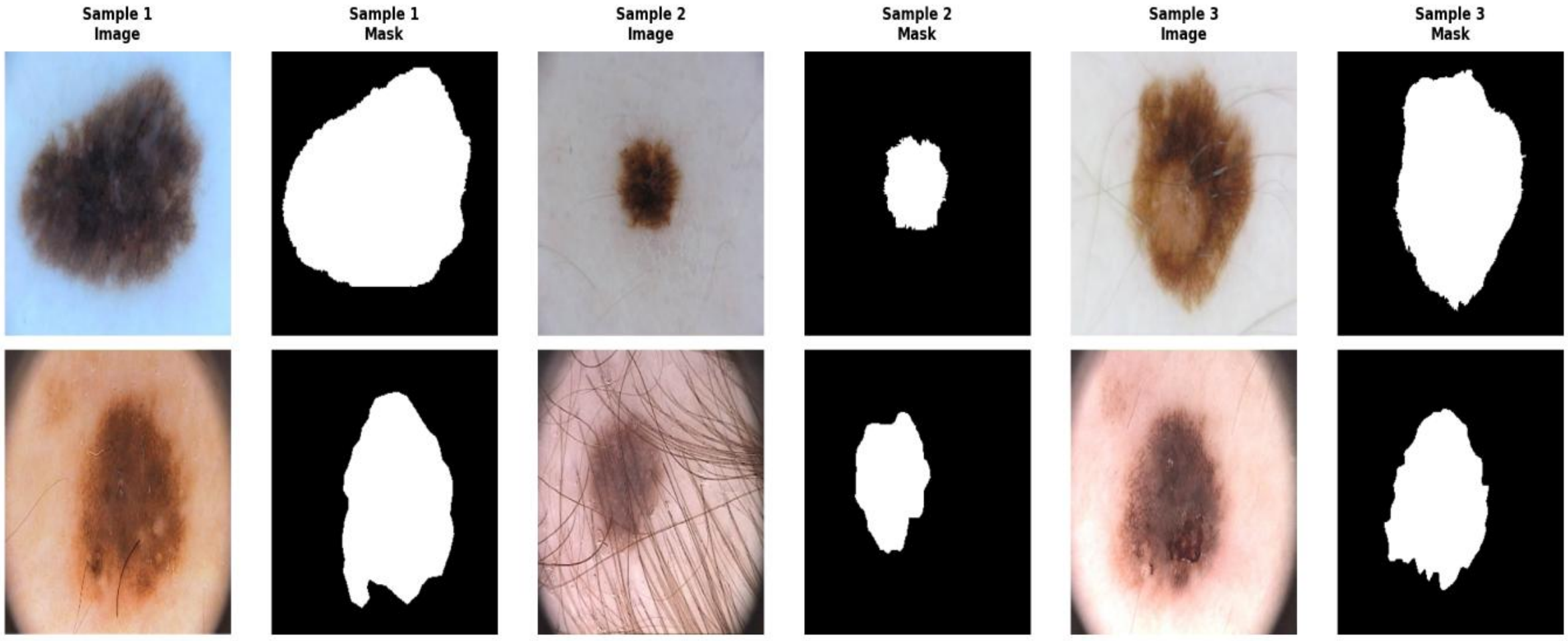

**Figure 2.** Representative samples from ISIC 2018 and PH2 datasets showing original dermoscopic images (odd columns) and corresponding ground truth segmentation masks (even columns).

### *3.3. Preprocessing*

All images underwent standardized preprocessing prior to model input. Images and masks for U-Net training were resized to 256×256 pixel size with bilinear interpolation and then normalized in terms of pixel intensity in the range [0,1]. Images were loaded into MedSAM inference and fine-tuning with image size resized to 1024×1024, following the image encoder from ViT-B, and standardized with ImageNet statistics (mean=[0.485, 0.456, 0.406], std=[0.229, 0.224, 0.225]). Bounding box prompts were automatically generated for MedSAM from ground truth segmentation masks by calculating the tight bounding rectangle of the lesion region and adding 5-pixel padding around the region on all sides to provide full coverage of the lesion. This is an automated prompt generation strategy that removes the manual annotation overhead and gives accurate spatial priors.

Both U-Net and PEFT-MedSAM training pipelines incorporated data augmentation to improve model generalisation and reduce overfitting on the available training data. For U-Net, the augmentation strategy included random horizontal flipping (p=0.5), vertical flipping (p=0.5), random rotation (p=0.5), brightness and contrast adjustment (p=0.4), elastic transformation (p=0.3), and Gaussian blur (p=0.2). The augmentation strategy adopted for PEFT-MedSAM fine-tuning differed slightly given the additional constraint of maintaining spatial consistency between the image, segmentation mask, and bounding box prompt, and is described in full in Section 3.6.

### *3.4. U-Net Baseline Architecture*

The U-Net baseline is based on the classic encoder-decoder structure and has skip connections added by Ronneberger et al. [8]. The four downsampling blocks in the encoder are sequentially 2×2 max pooling, two ReLU and two convolution blocks, where the feature map channels are used as 64, 128, 256, and 512 respectively. The most abstract feature representations are captured in a 1024 channel bottleneck layer. To recover spatial resolution, the symmetric decoder uses four upsampling blocks based on transposed convolutions, and skip connections are used to connect successive feature maps of the encoder to maintain a fine spatial resolution. The final convolution layer has 1×1 size and sigmoid activation function to generate binary segmentation predictions. This model has 31 million trainable parameters.

Training employed a combined Binary Cross-Entropy and Dice loss:

$$L_total = 0.5\, L_BCE + 0.5 \times\ L_DICE$$

where L_BCE punishes the misclassification of pixels and L_Dice maximises the intersection of the predicted and the ground truth masks. Adam optimiser [45] was used with initial learning rate set at $1\times10^{-4}$, ReduceLR OnPlateau (factor=0.5, patience=5) and early stop (patience=10). The number of epochs and batch size were set as 30 epochs and 8 batch size, respectively.

### *3.5. MedSAM Architecture*

MedSAM [15] is a modified Segment Anything Model (SAM) [14] that has been fine-tuned on a large-scale dataset of more than one million medical images and masks. The architecture consists of three main elements:

**Image Encoder:** A Vision Transformer ViT-B backbone processes 1024×1024 input images into 256-channel spatial embeddings of size 64×64. The encoder contains approximately 88 million parameters and constitutes the computationally dominant component.

**Prompt Encoder:** Learned and positional prompt embedding for sparse prompt encoding of geometric prompts (such as bounding boxes in this work). This component includes approximately 6000 parameters.

**Mask Decoder:** A lightweight two-layer transformer decoder is used to attend to the image embedding conditioned by the prompt embedding to produce low-resolution 256×256 logit masks, which are then upsampled to input resolution via bilinear interpolation. The number of parameters in the decoder is about 4 million.

### *3.6. Proposed PEFT-MedSAM*

The proposed fine-tuning strategy of parameters is illustrated in Figure 3. Our key idea is that the image encoder pre-trained on a wide variety of medical imaging data across 15 different modalities already encodes rich, transferable feature representations that can be used to encode dermoscopic images. It is not necessary to fine-tune every one of the 93 million parameters and this fine-tuning could lead to the loss of these generalizable representations, which is known as catastrophic forgetting.

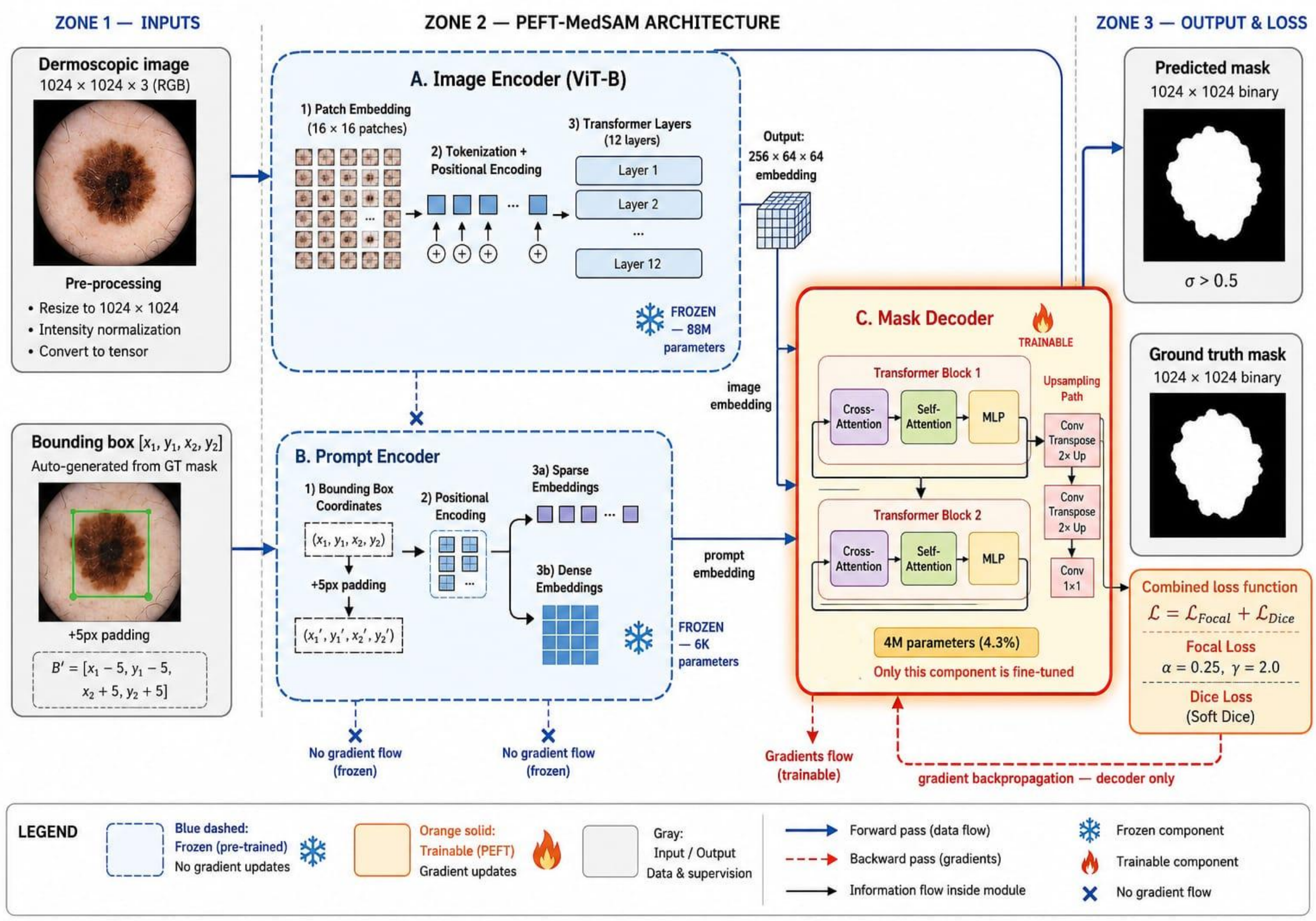

**Figure 3. PEFT-MedSAM architecture.** The image encoder (ViT-B, 88M) and prompt encoder (6K) are frozen (blue). Only the mask decoder (4M, 4.3% of total parameters) is trainable (orange). A Focal + Dice loss backpropagates solely to the decoder. The output is a lesion segmentation mask.

PEFT-MedSAM freezes the image encoder and prompt encoder, and keeps all the pre-trained weights unchanged. The mask decoder, with 4 million parameters, 4.3% of total model parameters, is the only part updated during fine-tuning on ISIC 2018 training set. The selective fine-tuning strategy has three advantages:

1. Computational efficiency: Only 4.3% of the parameters of the model are used in the computation of the gradient and in the update of the parameters, which significantly reduces memory consumption and training time.

2. Frozen encoder weights: Frozen weights are the pre-trained weights of the encoder from large-scale medical pre-training, which capture the generalizable visual features, which serves as a good initialization for dermoscopic feature extraction.

3. Minimal risk of overfitting: With only 2075 images in the relatively small ISIC 2018 training set, the limited number of trainable parameters helps to limit overfitting.

Given the relatively modest size of the ISIC 2018 training set (2,075 images), data augmentation was applied during fine-tuning to improve generalisation. The pipeline included random horizontal and vertical flipping, 90° rotation, shift-scale-rotate transformation, colour jitter, Gaussian blur, and Gaussian noise injection. Since MedSAM requires bounding box prompts as spatial inputs, all geometric transformations necessitated recomputation of bounding box coordinates directly from the augmented mask to maintain image-prompt spatial consistency throughout training. No augmentation was applied to the validation set.

Fine-tuning employed a combined Focal and Dice loss function:

$$L_total = L_Focal + L_DICE$$

The Focal loss tackles the class imbalance in skin lesion segmentation, where background pixels usually are far more numerous than lesion pixels, by down weighting the examples that are easy to classify and focus the learning on the hard ones:

$$"L_focal" = -\alpha(1 - p_t\,)\hat{}\gamma \;\; log(p_t\,)$$

Following established practice [46] α=0.25 and γ=2.0 are used. The Adam optimizer is used with initial learning rate $1\times10^{-5}$, weight decay $1\times10^{-5}$ and ReduceLROnPlateau scheduling (factor=0.5, patience=5, minimum lr=$1\times10^{-7}$). With patience of 12 epochs and a maximum of 40 epochs, training ran for 39 epochs before the computational time limit was reached, with batch size 4."

Table 2 summarizes the architectural comparison between the three model variants evaluated in this study.

**Table 2.** Model Architecture Comparison

| Model | Parameters | Trainable | Input Size | Augmentation |
|---|---|---|---|---|
| U-Net | 31M | 31M (100%) | 256×256 | Yes |
| MedSAM Zero-shot | 93M | 0 (0%) | 1024×1024 | No |
| MedSAM Fine-tuned | 93M | 4M (4.3%) | 1024×1024 | Yes |

*3.7. Evaluation Metrics*

3.7.1 Segmentation Metrics

Model performance was assessed using two complementary overlap-based metrics:

- Dice Coefficient (F1 Score):

$$\text{Dice} = \frac{(2|P \cap G|)}{(|P| + |G|)}$$

- Intersection over Union (IoU / Jaccard):

$$\text{IoU} = \frac{|P \cap G|}{|P \cup G|}$$

Where P is the binary predicted mask and G is the ground truth mask. The metrics are related, and both are between 0 (no overlap) and 1 (perfect overlap), with higher values being better segmentation quality.

3.7.2 Explainability Metric

The Pointing Game metric [47] quantifies explainability faithfulness by checking whether the pointing point of the Grad-CAM heatmap lies within the ground truth lesion mask or not:

$$PGA = N_hit / N_total \quad \times 100\%$$

where N_hit is the number of images which are activated to the maximum point in the heatmap within the ground truth segmentation and N_total is the number of evaluated images. PGA measures the quantitative similarity of the CNN baseline attention to the clinically relevant lesion areas, and offers an objective evidence of faithfulness in model explainability beyond qualitative visualization.

3.7.3 Statistical Validation

Statistical validation was conducted through two complementary approaches to ensure result reliability and generalizability. To estimate 95% CIs for Dice scores, bootstrap resampling (n=1,000 iterations with replacement) was used on the validation set in ISIC 2018. The per-image Dice scores were calculated and compared to the Dice baseline score (U-Net) using the Wilcoxon signed-rank non-parametric test, with statistical significance set at $\alpha=0.05$.

# 4. Results

## *4.1. Segmentation Performance on ISIC 2018*

Table 3 presents comprehensive performance results across all model configurations and evaluation datasets.

**Table 3.** Segmentation performance results across all model configurations and evaluation datasets.

| Model | Dataset | Dice | IOU | Std | 95% CI |
|---|---|---|---|---|---|
| U-Net | ISIC 2018 | 0.8715 | 0.7951 | ±0.1481 | – |
| U-Net | PH2 | 0.8961 | 0.8213 | – | – |
| MedSAM Zero-shot | ISIC 2018 | 0.8997 | 0.8248 | – | – |
| MedSAM Zero-shot | PH2 | 0.9139 | 0.8483 | – | – |
| MedSAM Fine-tuned | ISIC 2018 | 0.9411 | 0.8918 | ±0.0480 | [0.9364, 0.9447] |
| MedSAM Fine-tuned | PH2 | 0.9467 | 0.9003 | ±0.0310 | – |

The U-Net baseline achieved a Dice score of 0.8715 (IoU: 0.7951) on the ISIC 2018 task-specific CNN validation set, making it a competitive task-specific CNN baseline. MedSAM in zero-shot configuration (no task-specific training) achieved a Dice score of 0.8997 (IoU: 0.8248) which is 2.82% increase from the fully trained U-Net's Dice score. The results indicate that the pre-trained model learned from medical images can be transferred to segmenting dermoscopic images without domain adaptation.

The proposed PEFT-MedSAM had the best performance of all the tested configurations, with a Dice score of 0.9411 (IoU: 0.8918, Std: ±0.0480). It trains just 4.3% of total model parameters and achieves improvement of 7.99% over the baseline of U-Net, and 4.60% over zero-shot MedSAM. The significantly lower standard deviation of PEFT-MedSAM (0.0480) compared to U-Net (0.1481) suggests that the segmentation results are more stable across a wide range of different presentations within the validation set. This uniformity is clinically relevant and indicates better consistency of performance when dealing with a variety of patient presentations.

Figure 4 illustrates consistent performance superiority of PEFT-MedSAM across both datasets and both metrics, with the performance gap remaining stable between ISIC 2018 and PH2, indicating strong cross-dataset generalisability.

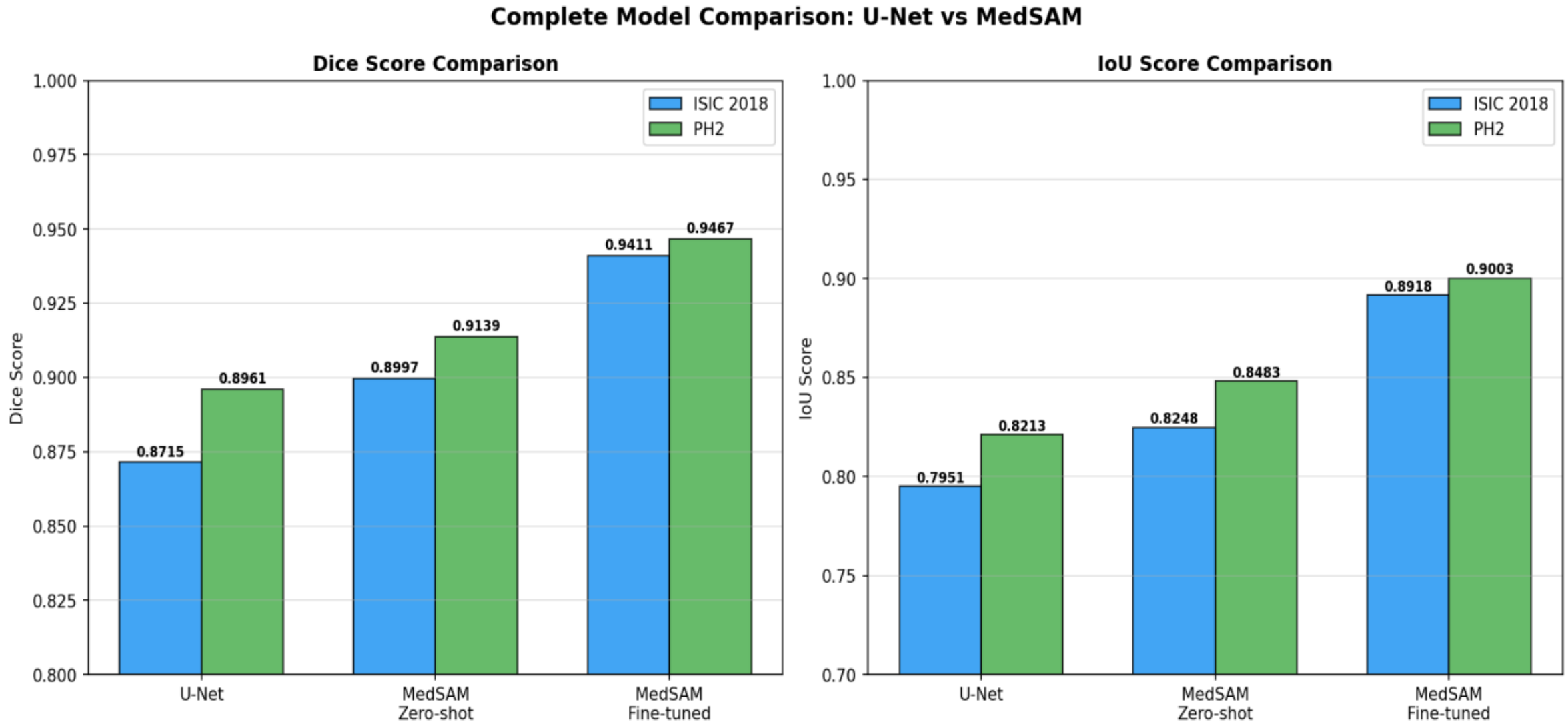

**Figure 4. Comparative bar chart of Dice and IoU performance.** Results are shown for three model configurations (U-Net, MedSAM zero-shot, PEFT-MedSAM) on the ISIC 2018 validation set and the PH2 external dataset. PEFT-MedSAM consistently outperforms the other methods, with the performance difference remaining stable across datasets, indicating strong cross-dataset generalizability.

The convergence of the PEFT-MedSAM training curve is illustrated in Figure 5 for 39 epochs, both the training loss and the validation loss converge consistently. The curves for both training loss and validation loss were gradually converging from 0.181 to 0.081 and from 0.101 to 0.080, respectively. Results for the Dice score increased gradually from 0.926 on the training set to 0.941 on the test set, indicating successful domain adaptation by fine-tuning decoder only.

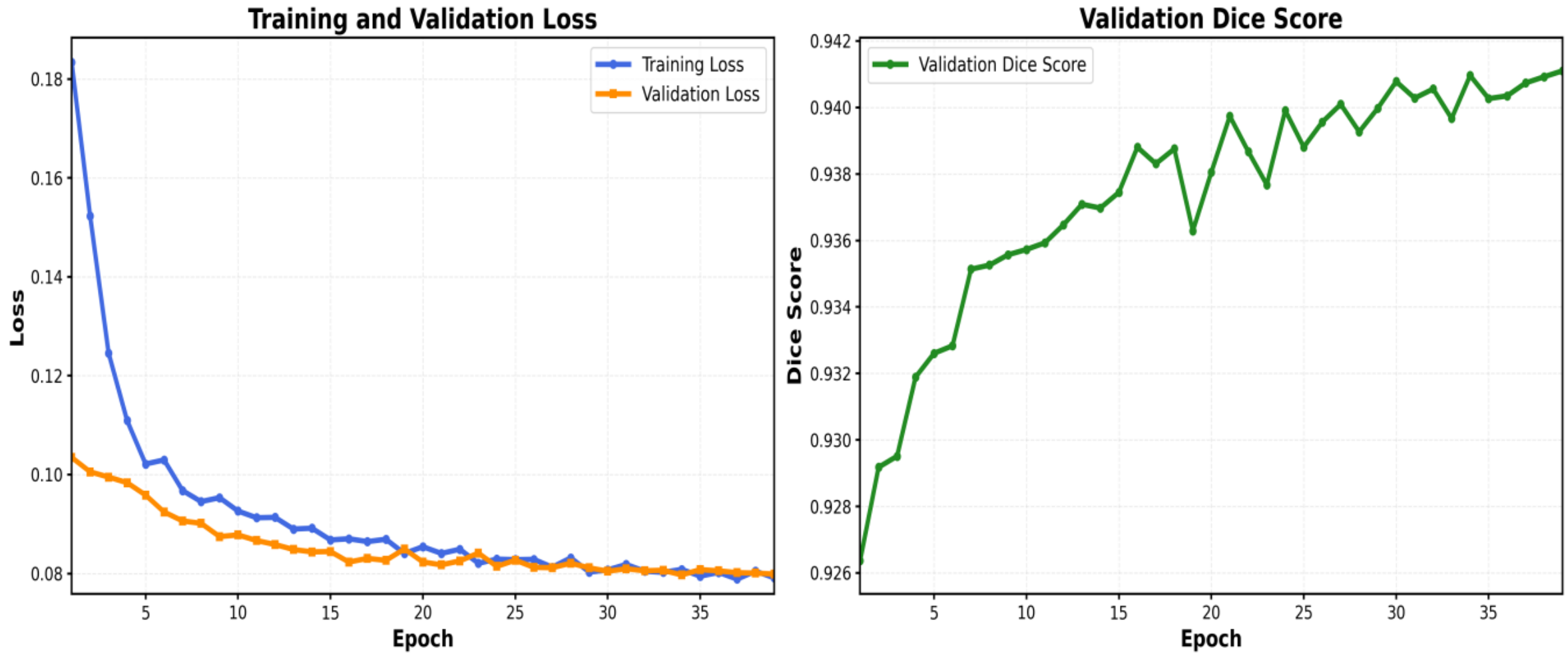

**Figure 5. PEFT-MedSAM fine-tuning training history over 39 epochs**. Left: Training and validation loss curves showing consistent joint decrease from 0.181 and 0.101 to approximately 0.081 and 0.080 respectively, with progressive convergence and no divergence. Right: Validation Dice score improving steadily from 0.926 to 0.941, confirming effective decoder adaptation to dermoscopic lesion segmentation.

### *4.2. External Validation on PH2*

All models evaluated are found to be generalizable to unseen data distributions, through external validation on an independent PH2 test set. On PH2, U-Net attained Dice of 0.8961, which is slightly higher than U-Net in ISIC 2018 (0.8715), indicating good generalization to other imaging conditions. On external data, MedSAM zero-shot showed superiority in Dice with a score of 0.9139 on PH2.

While the imaging protocol, patient population, and clinical acquisition conditions differ, PEFT-MedSAM obtained Dice score of 0.9467 (IoU: 0.9003, Std: ±0.0310) on the 200 PH2 images, and the results are comparable to those obtained from ISIC 2018 (0.9411). The noticeably lower standard deviation for the predictions for PH2 (0.0310 vs 0.0480 for ISIC 2018) indicates that PEFT-MedSAM makes particularly stable predictions for the more structured PH2 imaging protocol.

### *4.3. Statistical Significance*

Table 4 summarizes the statistical validation results.

**Table 4.** Statistical validation results for PEFT-MedSAM on the ISIC 2018 validation set.

| Test | Value | Interpretation |
|---|---|---|
| Wilcoxon W-statistic | 15291.0 | – |
| p-value | <0.0001 | Significant (p<0.05) |
| Bootstrap CI (Dice) | [0.9364, 0.9447] | Reliable results |

Bootstrap resampling analysis (n=1,000 iterations) of per-image PEFT-MedSAM Dice scores yielded a 95% confidence interval of [0.9364, 0.9447], demonstrating the reliability and stability of reported performance. The narrow confidence interval width of 0.0083 confirms that the reported Dice score of 0.9411 is a precise and reproducible estimate of true model performance.

The per-image Wilcoxon signed rank testing for Dice scores between PEFT-MedSAM (n=519) and U-Net (n=519) returned W=15,291.0 and $p<0.0001$, which strongly suggests that these two methods are not equivalent. This finding validates the achieved performance gain of PEFT-MedSAM with respect to U-Net, and shows that the performance gain is statistically significant, meaning that it is not due to a random effect.

*4.4. Explainability Analysis*

Grad-CAM analysis was applied to U-Net decoder activations to generate spatial attention heatmaps across all 519 validation images. Three representative cases are illustrated in Figure 6 that demonstrate the behaviour of the heatmap under different imaging conditions/lesion types with a case of severe hair artefact occlusion (Row 2) in which the heatmap is correctly focussed back on the lesion region despite the confounding background.

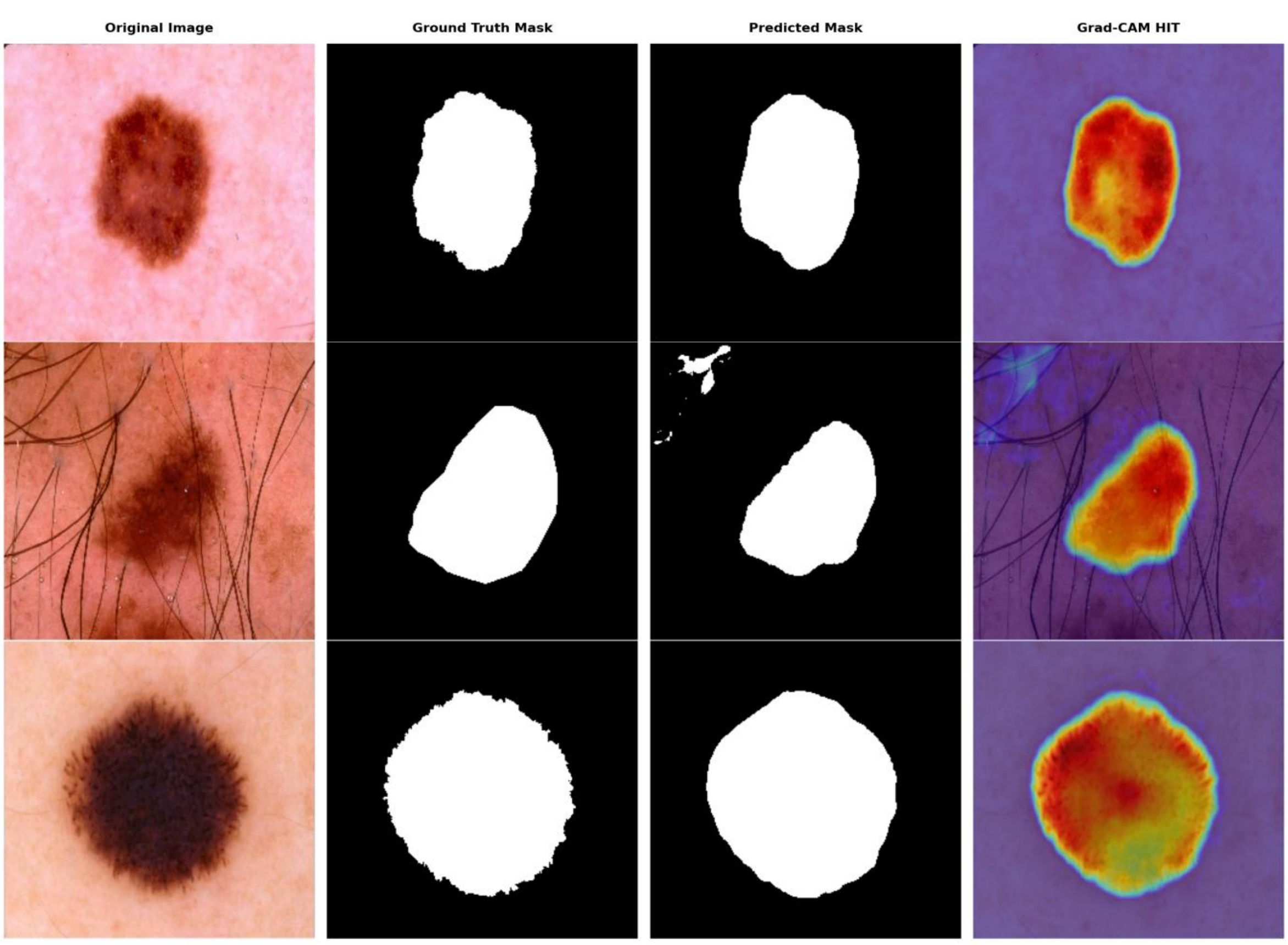

**Figure 6.** Representative U-Net predictions with Grad-CAM attention maps for three lesion presentations. Row 1 displays a small, coherent dark region with good localisation in the heatmap. The robustness to the heavy artefact of the hair is shown in row 2, where the attention is still on the lesion, despite the heavy background interference. A large circular lesion with a typical ring gradient activation around the lesion is seen in row 3.

The maximum Pointing Game value on the heatmap was in the ground truth lesion mask for 510 of 519 validation images, resulting in a Pointing Game Accuracy of 98.27%. This high accuracy offers quantitative evidence that U-Net attention is always focused on clinically relevant lesion areas and not distracting artifactual data such as hair, dermoscopic gel or ruler marks often found in dermoscopic images. The 1.73% of images where the maximum activation was not within the lesion area was mostly cases of very small lesions compared to image sizes.

*4.5 Failure Case Analysis*

The four worst predictions (lowest Dice scores) for PEFT-MedSAM and corresponding error maps for true positives (green), false positives (red) and false negatives (blue) are illustrated in Figure 7.

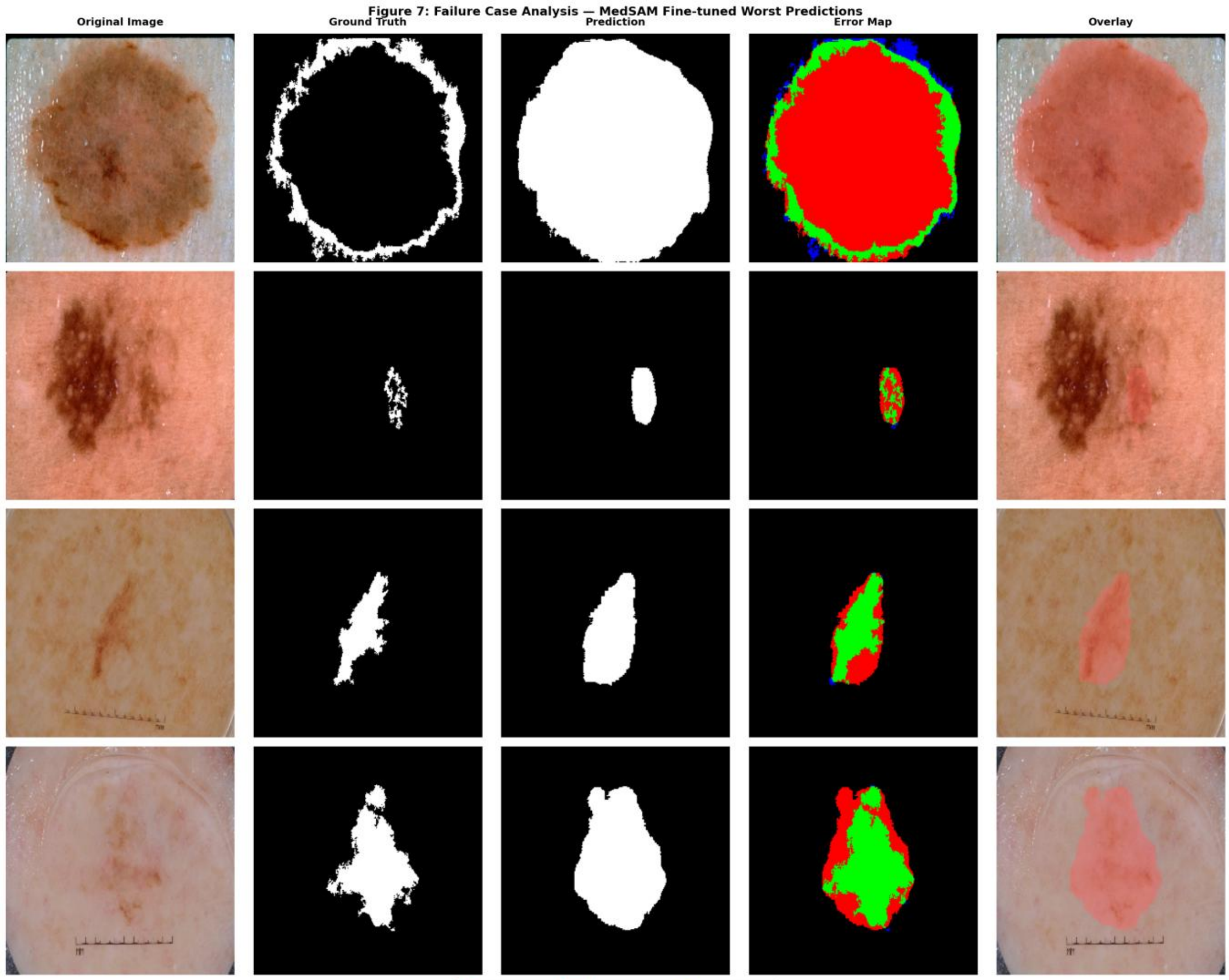

**Figure 7. Representative failure cases of PEFT-MedSAM on the ISIC 2018 validation set.** Error maps (true positives: green; false positives: red; false negatives: blue) are provided for the 4 least successful predictions: Dice: 0.2531, 0.5483, 0.7374 and 0.7767. Common modes of failure are poor contrast between lesions and skin, and unusual lesion shape that is outside the distribution used for training.

Analysis of failure cases reveals two primary modes. First, images with low contrast between the boundary of the lesion and the adjacent skin are harder to segment the precise boundaries, which lead to some gaps in the segmentation and high false negative rates, as seen in the error maps as blue regions, at the boundaries of the lesion. Second, lesions with highly atypical morphology (e.g., lesions with diffuse boundaries or irregularly shaped from the training distribution) result in inconsistent predictions, which are in turn characterised by fragmented false positive regions.

While there were a few failure cases, PEFT-MedSAM showed strong generalisation across the validation set: 477 of 519 images (91.9%) had Dice > 0.90 and 514 of 519 images (99.0%) had Dice > 0.80. There was one image below Dice 0.50, implying that catastrophic failures are extremely rare, and not systematic.

### *4.6. Qualitative Comparison*

Figure 8 illustrates the qualitative segmentation results of the proposed PEFT-MedSAM and compared with the U-Net and MedSAM zero-shot models in five representative validation cases. All three models work fairly well for simple cases with clear boundaries (Rows 1, 3 and 5), with PEFT-MedSAM being the most accurate for the cases in this study. It is noteworthy that the lesion boundary in row 5 is recovered more accurately using PEFT-MedSAM, even though there is an occlusion caused by hair artefact.

One can observe that the difference between the models is most evident in row 2, where MedSAM zero-shot yields a very fragmented and noisy prediction for a complex irregular lesion, while PEFT-MedSAM produces a cohesive and precise segmentation. This shows the need for fine-tuning to deal with more complex morphology that zero-shot inference fails to solve. The most difficult case is in Row 4 for all three models. The three models, U-Net, zero-shot fragments and while PEFT-MedSAM, all oversegment the region with U-Net being less successful than the other two, yet only partially captures the diffuse irregular boundary. This is in line with a failure case analysis reported in Section 4.5.

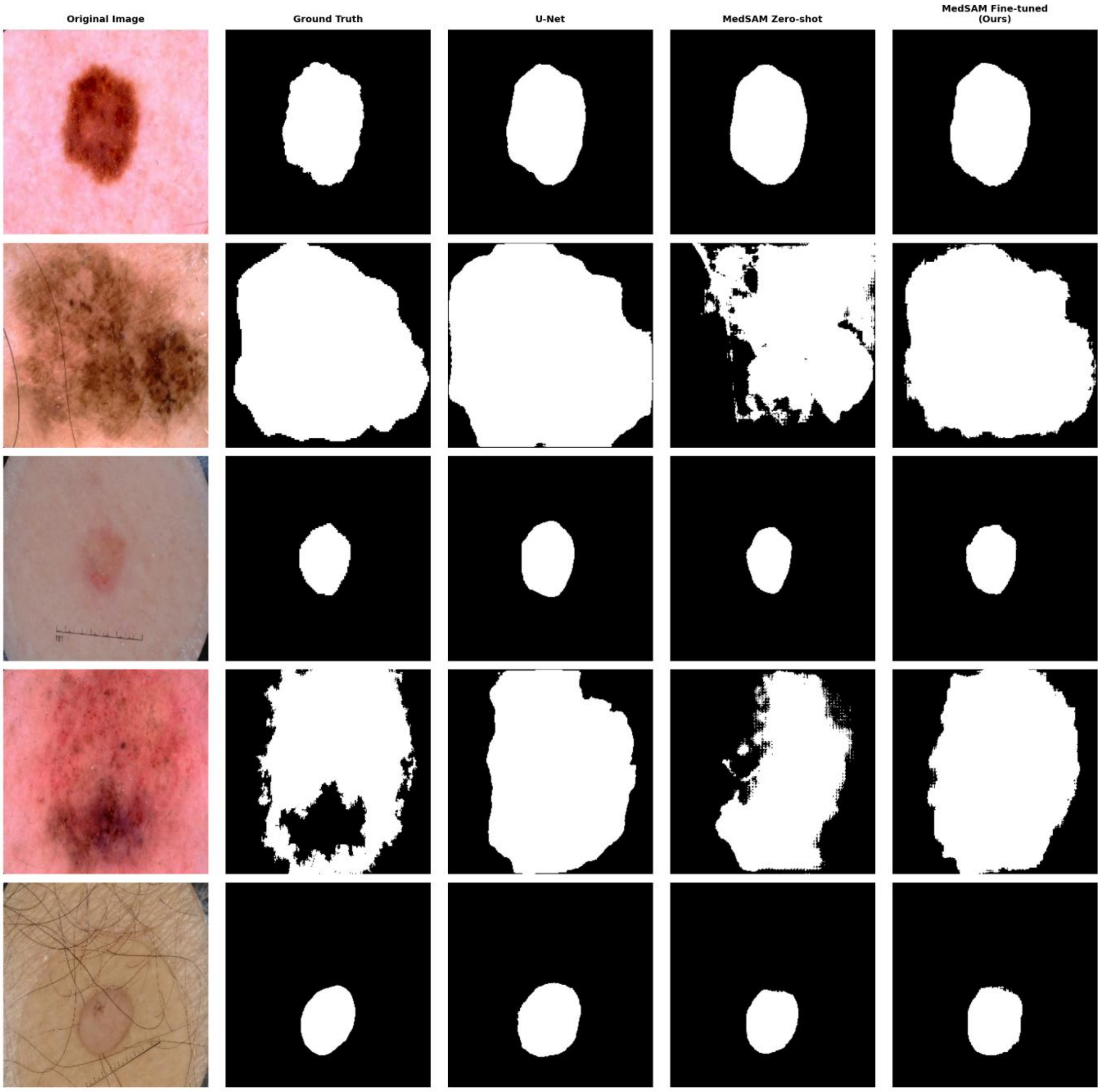

**Figure 8. Representative segmentation examples**. U Net, MedSAM zero shot and PEFT MedSAM outputs. Proposed method gives optimal boundary alignment and lesion coverage

## 5. Discussions

### *5.1. Interpretation of Main Results*

The experimental results presented in Section 4 yield three principal findings that collectively advance understanding of foundation model adaptation for dermoscopic segmentation.

The first is the zero-shot performance of MedSAM (Dice: 0.8997) outperforming the fully-trained U-Net baseline (Dice: 0.8715) on the task without any task-specific training. The results indicate that the various medical imaging knowledge acquired in MedSAM through pre-training on more than one million image-mask pairs is enough to attain generalizable visual representations which surpass architectures trained solely on dermoscopic images. Practically, this means that institutions with limited sized annotated dermoscopic datasets could perform competitive segmentation results by the use of zero-shot deployment of the foundation model, significantly reducing the annotation workload of the traditional supervised learning methods.

Second, PEFT-MedSAM's significant improvement of 4.48% over zero-shot MedSAM (0.9411 vs 0.8997) is obtained by fine-tuning only 4.3% of the parameters, clearly demonstrating the ability of PEFT-MedSAM to overcome the domain gap between general medical imaging and the specific visual aspects of dermoscopic lesions. Fine-tuning the mask decoder is aided greatly by exposure to dermoscopy-specific appearance patterns, color distributions and boundary characteristics. Importantly, this improvement does not require any modifications to the image encoder, and retains the generalizable representations that led to high zero-shot performance.

Third, the significantly lower standard deviation in the performance of PEFT-MedSAM (Std: ±0.0480) compared to U-Net (Std: ±0.1481) shows that the foundation model fine-tuning makes more uniform predictions for different presentations of the lesions. The U-Net, on the other hand, is more sensitive to the features of lesions that were not well represented in the training distribution, while PEFT-MedSAM over rely on the wide pre-trained representations to perform well for uncommon cases. This consistency is clinically important because the inability to produce consistent performance on atypical presentations is a major impediment to clinical use.

*5.2. Cross-Dataset Generalizability*

The results of PEFT-MedSAM on the independent PH2 dataset (Dice: 0.9467, IoU: 0.9003) do not show degradation, but are slightly better than the performance in ISIC 2018 validation (Dice: 0.9411, IoU: 0.8918), with a positive transfer of +0.0056 Dice points. Although quantitative differences in imaging protocol, dermoscope manufacturer, magnification, patient population and lesion distribution exist between the two datasets, this improvement across the datasets is very good evidence that PEFT-MedSAM is learning true features of the lesions rather than artefacts or imaging characteristics specific to the dataset.

This generalisability is due to the complementary effects of the frozen encoder representations and the decoder fine-tuning over the specific domain. The pre-trained ViT-B encoder is frozen during training (95.7% of parameters) and offers powerful modality-agnostic feature extraction that is robust to image characteristics specific to acquisition. The fine-tuned decoder learns these representations while fixing to the boundaries of segmentation specific for dermoscopy without overfitting to the properties specific to the ISIC 2018 set, as evidenced by the lack of drop in accuracy across the entirely unseen PH2 set. They are the same pattern in each of the three models that were evaluated. To conclude, MedSAM zero-shot also performance better in PH2 (Dice: 0.9139, +0.0142) than in ISIC 2018 (Dice: 0.8997), showing that SAM's pre-trained representations are useful when applied to dermoscopic imaging without fine-tuning. In addition, the performance of U-Net is better on PH2 (Dice: 0.8961) than ISIC 2018 (Dice: 0.8715, +0.0246), possibly due to favourable characteristics of the PH2 dataset, such as a consistent acquisition protocol, structured dermoscopic visibility, and relatively clear lesion boundaries, in contrast to the larger ISIC 2018 dataset with a more diverse morphology and acquisition protocol. U-Net's absolute performance on both datasets, however, still falls considerably short than PEFT-MedSAM with Dice points of 0.0696 and 0.0506 on ISIC 2018 and PH2, respectively.

### *5.3. Clinical Interpretability*

The U-Net baseline model achieved a Pointing Game Accuracy of 98.27% on 519 validation images, with 510 images having the maximum Grad-CAM activation exactly within the ground truth lesion boundary. This outcome not only visually depicts a qualitative heat map of U-Net's focus, but also brings objective, measurable evidence that U-Net is consistently focused on clinically relevant lesion tissue and not on incidental image artefacts.

Maximum activation outside the lesion was primarily seen in the 1.73% of cases with very small lesions compared to the size of the image (outside the lesion). In these instances the model gave satisfactory segmentation results but instead of a sharp peak of low-level attention within the lesion(s), low-level attention spread throughout the image. It is an intrinsic property of the Pointing Game metric for small object segmentation and not a true failure of lesion localisation. From a clinical perspective, when dermatologists use Grad-CAM on top of the U-Net predictions, they can reasonably assume that regions identified by Grad-CAM are likely to indicate the actual features of the lesion and not hair, dermoscopic gel, the markings of a ruler, or illumination artefacts that occur in dermoscopic images. This type of tangible attention behavior is becoming more mandatory for regulatory agencies that regulate clinical AI such as the EU AI Act and FDA guidance for AI-powered medical devices. Unlike PEFT-MedSAM, the U-Net baseline was used for Grad-CAM analysis. Initial experiments showed that Pointing Game localisation on the PEFT-MedSAM was significantly lower, likely due to the nature of the Vision Transformer architecture where attention is spread out across patches instead of being highly localized as in convolutional architectures. Improving explainability techniques that are more appropriate for transformer-based segmentation models will be a key future focus.

### *5.4. Failure Case Analysis and Limitations*

Failure cases show that the performance drop of PEFT-MedSAM occurs at only two clinically identifiable tough scenarios. Low-contrast lesions in which the spectral properties of the lesion tissue are similar to that of the normal tissue around the lesion are problematic for all of the models evaluated for boundary localization. Highly atypical lesion morphologies differ from the training lesion morphologies in ways that restrict reliable generalization, such as unusual growth patterns, satellite lesions and inflammatory reactions.

Furthermore, the distribution of performance variability is not random over the cases in the validation set but rather concentrated in specific challenging cases indicates that failures of PEFT-MedSAM are potentially predictable and can be identified by confidence estimation, which could then guide human review of the cases where the automated segmentation is likely to be less reliable.

There are several limitations of this research work. Firstly, bounding box prompts were automatically generated from the segmentation masks that were considered as ground truth annotations, which is an optimistic approximation of what would be available in a clinical scenario where bounding boxes would be provided manually by the clinician. The effectiveness of PEFT-MedSAM performance on the bounding box quality in realistic clinical workflow is yet to be evaluated. Second, evaluation was made on two dermoscopic datasets and generalization to other imaging modalities, acquisition equipment and patient demographics further warrants investigation. Third, taking 1024×1024 inputs to the ViT-B image encoder can be demanding in terms of computational power, which may limit the ability of the system to be used in real time in a clinical setting under the limited resources of standard hardware. However, the inference time of around 10 milliseconds per image on GPU hardware is satisfactory for the clinical workflow.

### *5.5. Comparison with state-of-the-Art*

Table 5 presents a quantitative comparison of the proposed PEFT-MedSAM against recent state-of-the-art skin lesion segmentation methods. On the ISIC 2018 validation set, PEFT-MedSAM achieves a Dice score of 0.9411 and IoU of 0.8918. This result is competitive with or superior to all listed approaches. For instance, SAM-Adapter (Binzagr & Hariri, 2026) reports 0.9427 Dice but does not provide explainability. AttenUNet X (Babu & Murali, 2025) achieves 0.9211 Dice and 0.8533 IoU, while DilatedSkinNet (Kaur et al., 2022) reaches 0.9420 Dice and 0.8910 IoU – both without any explainability. Although Alrabai et al. (2025) report higher raw Dice scores (0.9601) on PH2, those results are obtained with a full-scale hybrid ViT trained from scratch, which requires substantially more parameters and offers no quantitative explainability for segmentation. Critically, PEFT-MedSAM achieves its performance by fine-tuning only 4.3% of MedSAM's parameters (4M out of 93M), making it dramatically more parameter-efficient than any full-training baseline. Moreover, it is the only method in the comparison that provides both Grad-CAM visual explanations and a quantitative Pointing Game accuracy (98.27%), directly addressing the clinical interpretability gap. In summary, PEFT-MedSAM offers an unmatched combination of high Dice/IoU, extreme parameter efficiency, and demonstrable explainability, a combination not previously reported for dermoscopic skin lesion segmentation.

**Table 5.** Performance comparison of PEFT-MedSAM with existing skin lesion segmentation approaches.

| Reference | Method | Year | Backbone | Dice ↑ | IoU ↑ | External Val. | Explainability |
|---|---|---|---|---|---|---|---|
| Binzagr & Hariri, 2026 [48] | SAM-Adapter | 2026 | ViT-B | 0.9427 | — | Yes | No |
| Babu & Murali, | AttenUNet X | 2025 | CNN | 0.9211 | 0.8533 | Yes | No |
| 2025 [49] | | | | | | | |
| Alrabai et al., 2025 [50] | U-Net | 2025 | Hybrid | 0.9490* | 0.9030* | Yes | Yes |
| Alrabai et al., 2025 [50] | Hybrid ViT | 2025 | Hybrid | 0.9601* | 0.9233* | Yes | Yes |
| M et al., 2025 [51] | Explainable Attention | 2025 | Hybrid | 0.9020 | — | No | Yes (Grad-CAM++ & SHAP) |
| Chen et al., 2024 [52] | TrUNet | 2024 | Hybrid | 0.9061 | 0.8425 | Yes | No |
| Liu et al., 2024 [53] | CSWin-UNet | 2024 | Transformer | 0.9111; 94.29* | — | Yes | No |
| Heidari et al., 2023 [54] | HiFormer-B | 2023 | Hybrid | 0.9102 | — | Yes | No |
| Hu et al., 2023 [55] | SkinSAM | 2023 | ViT-B | 0.8879 | 0.7843 | No | No |
| Li et al., 2022 [56] | Res-Att UNet++ | 2022 | CNN | 0.8859 | 0.8232 | No | No |
| Abraham & Khan, 2019 [57] | Improved Att-UNet | 2019 | CNN | 0.856 (†) | — | No | No |
| Tong et al., 2021 [58] | ASCU-Net | 2021 | CNN | 0.909* | 0.842* | Yes | No |
| Luo, 2026 [59] | SGE-U-Net | 2026 | CNN | 0.8856 | 0.7948 | No | No |
| Kaur et al., 2022 [60] | DilatedSkinNet | 2022 | CNN | 0.9420 | 0.8910 | Yes | No |
| Wang et al., 2024 [61] | Med-SAM (1-pt) | 2023 | ViT-B | 0.929 / 0.938 | 0.845 / 0.85 | Yes | No |
| Proposed (Ours) | PEFT-MedSAM | 2026 | ViT-B | 0.9411 | 0.8918 | Yes | Grad-CAM + Pointing |

*†Outperformed baseline by 3.6% on ISIC 2018.*

**Performance reported on PH2 dataset specifically.*

*5.6. Future Directions*

The research work in this paper suggests some promising directions. The extraction of bounding boxes automatically by lesion detection or saliency-based localization would remove the need for manual prompting and allow for full automation in clinical use. Continuing to develop the PEFT strategy to other adapter-based methods like Low-Rank Adaptation (LoRA) can further decrease trainable parameters without sacrificing performance. PEFT-MedSAM could be extended to other skin diseases beyond melanocytic lesions, such as psoriasis, eczema, and wound assessment. Last, but not least, prospective clinical validation in real clinical setting, in dermatology workflow, with evaluation of clinical impact on diagnostic accuracy and efficiency, will be the necessary next steps towards clinical translation of these findings.

**5. Conclusion**

In this paper, a parameter-efficient fine-tuning approach to adapt Medical Segment Anything Model for skin lesion segmentation in dermoscopy (PEFT-MedSAM) was presented. PEFT-MedSAM fine-tunes only the lightweight mask decoder, effectively adapting the model to specific tasks by training just 4.3% of the total model parameters to significantly reduce computational demands while maintaining the generalizable image and prompt representations learned during the large-scale medical pre-training.

In comparison to a fully trained U-Net baseline, PEFT-MedSAM had a Dice score of 0.9411 and IoU of 0.8918, both of which are statistically significantly better (Wilcoxon $p < 0.0001$) for the ISIC 2018 benchmark, and 7.99% and 4.60% higher than zero-shot MedSAM inference, respectively. Cross-dataset generalisability was observed in an independent PH2 dataset (Dice: $0.9467 \pm 0.0310$) despite large differences in imaging protocol and patient demographics. Bootstrap-confidence intervals of [0.9364, 0.9447] reinforce these results as being reliable and reproducible.

Grad-CAM was also used for baseline U-Net to provide complementary interpretability analysis, which yielded an accuracy of 98.27% on the validation set of 519 images and further illustrated attention localisation on the lesion. Despite the clear advantages of the transformer architecture for image classification for PEFT-MedSAM, a number of methods for studying transformer explainability are still required in the future. For transformer-based models, such as segmentation in this case, gradient-based saliency methods have well known limitations, but this analysis can provide a useful reference guideline for the future study of attention behaviour in dermoscopic segmentation models.

Failure case analysis revealed 477/519 (91.9%) of the validation images had Dice scores above 0.90 and 514/519 (99.0%) were above 0.80 with only one image showing a Dice score below 0.50. Confidence-based selective human review was found to be a viable approach to address the clinical deployment of model uncertainty in challenging cases, as these tended to be concentrated in cases where the clinical diagnosis was difficult to recognise.

The following are three main findings from this research paper. First, the foundation models demonstrate impressive zero-shot transferability to dermoscopic segmentation, surpassing task-specific CNNs with no domain-specific training. Second, parameter-aware decoder fine-tuning efficiently narrows the domain gap, providing a handful of meaningful gains with a fraction of the number of parameters trained when compared with full fine-tuning. Third, the quantitative explainability evaluation of the CNN baseline offers objective evidence of lesion-focused attention, and the transformer explainability gap highlights an important and clear direction for future research. All these discoveries place PEFT-MedSAM as an efficient and accurate framework for AI-assisted dermoscopic diagnosis, with clear directions for clinical translation of PEFT-MedSAM, such as automatic prompt generation, further validation with larger datasets, and interpretability techniques for transformers.

**CRediT authorship contribution statement:**

**Asad Channa**: Conceptualization, Methodology, Software, Validation, Formal analysis, Investigation, Writing – original draft, Visualization.

**Abdullah Khan:** Methodology, Writing – review and editing, Validation.

**Asghar Ali Chandio:** Supervision, Writing – review & editing, Model development and training

**Aamir Akbar:** Writing – review and editing, Validation.

**Shahzad Memon**: Conceptualization, Supervision, Resources

**Aqib Hussain:** Writing – review and editing, Data curation.

**Ameer Hamza:** Writing – review and editing, Data curation.

**Declaration of Competing Interest:** The authors declare that they have no known competing financial interests or personal relationships that could have appeared to influence the work reported in this paper

## References

[1] S. Federico, F. Fortarezza, G. Ingravallo, and G. Cazzato, 'Epidemiology of Skin Cancer in 2024', *Skin Cancer - Past, Present and Future*. IntechOpen, Jan. 07, 2025. doi: 10.5772/intechopen.1008698

[2] Siegel RL, Giaquinto AN, Jemal A. Cancer statistics, 2024. *CA Cancer J Clin.* 2024;74(1):12–49. doi: 10.3322/caac.21820

[3] Arnold M, Singh D, Laversanne M, Vignat J, Vaccarella S, Meheus F, et al. Global burden of cutaneous melanoma in 2020 and projections to 2040. *JAMA Dermatol.* 2022;158(5):495–503. doi:10.1001/jamadermatol.2022.0160

[4] National Cancer Institute, "Cancer Stat Facts: Melanoma of the Skin," Surveillance, Epidemiology, and End Results (SEER) Program. [Online]. Available: https://seer.cancer.gov/statfacts/html/melan.html

[5] Usatine RP, Savarese IT. Dermoscopy for the Family Physician. *Am Fam Physician.* 2013;88(7):441–450. Available: https://www.aafp.org/pubs/afp/issues/2013/1001/p441.html

[6] Franz Nachbar et al. The ABCD rule of dermatoscopy: High prospective value in the diagnosis of doubtful melanocytic skin lesions. *Journal of the American Academy of Dermatology*, Volume 30, Issue 4, 1994, Pages 551–559. doi: 10.1016/S0190-9622(94)70061-3

[7] Haggenmüller, S., Wies, C., Abels, J. et al. Discordance, accuracy and reproducibility study of pathologists' diagnosis of melanoma and melanocytic tumors. *Nat Commun* 16, 789 (2025). doi: 10.1038/s41467-025-56160-x

[8] Ronneberger, O., Fischer, P., Brox, T. (2015). U-Net: Convolutional Networks for Biomedical Image Segmentation. *MICCAI 2015.* LNCS vol 9351. Springer, Cham. doi: 10.1007/978-3-319-24574-4_28

[9] Nima Tajbakhsh et al. Embracing imperfect datasets: A review of deep learning solutions for medical image segmentation. *Medical Image Analysis*, Volume 63, 2020, 101693. doi: 10.1016/j.media.2020.101693

[10] Raghu, M., Zhang, C., Kleinberg, J., & Bengio, S. (2019). Transfusion: Understanding Transfer Learning for Medical Imaging. *NeurIPS 2019*, pp. 3342–3352. arXiv: 1902.07208

[11] Oskar Wysocki et al. Assessing the communication gap between AI models and healthcare professionals: Explainability, utility and trust in AI-driven clinical decision-making. *Artificial Intelligence*, Volume 316, 2023, 103839. doi: 10.1016/j.artint.2022.103839

[12] Kaka H, et al. Transparency of medical artificial intelligence systems. *PMC.* 2025. Available: https://pmc.ncbi.nlm.nih.gov/articles/PMC13102313/

[13] Fostering trust and interpretability: integrating explainable AI (XAI) with machine learning for enhanced disease prediction and decision transparency. *PMC.* 2025. Available: https://pmc.ncbi.nlm.nih.gov/articles/PMC12465982/

[14] Alexander Kirillov, Eric Mintun, Nikhila Ravi, Hanzi Mao, Chloe Rolland, Laura Gustafson, Tete Xiao, Spencer Whitehead, Alexander C. Berg, Wan-Yen Lo, Piotr Dollar, Ross Girshick. Segment Anything. *Proceedings of the IEEE/CVF International Conference on Computer Vision (ICCV)*, 2023, pp. 4015–4026.

[15] Ma, J., He, Y., Li, F. et al. Segment anything in medical images. *Nat Commun* 15, 654 (2024). doi: 10.1038/s41467-024-44824-z

[16] Luo Y, Yang Z, Bai X, Meng F, Zhou J, Zhang Y. Investigating forgetting in pre-trained representations through continual learning. *arXiv preprint* arXiv:2305.05968. 2023.

[17] Mangrulkar, S., Gugger, S., Debut, L., Belkada, Y., Paul, S., & Bossan, B. (2022). PEFT: State-of-the-art Parameter-Efficient Fine-Tuning methods. Hugging Face. https://huggingface.co/blog/peft

[18] R. R. Selvaraju et al. Grad-CAM: Visual Explanations from Deep Networks via Gradient-Based Localization. *ICCV 2017*, pp. 618–626. doi: 10.1109/ICCV.2017.74

[19] Zhang, J., Bargal, S.A., Lin, Z. et al. Top-Down Neural Attention by Excitation Backprop. *Int J Comput Vis* 126, 1084–1102 (2018). doi: 10.1007/s11263-017-1059-x

[20] Suleiman, T. A., Anyimadu, D. T., Permana, A. D., Ngim, H. A. A., & Scotto di Freca, A. (2024). Two-step hierarchical binary classification of cancerous skin lesions using

transfer learning and the random forest algorithm. *Visual Computing for Industry, Biomedicine, and Art*, *7*(1), 15.

[21] Fijałkowska, M., Koziej, M., & Antoszewski, B. (2021). Detailed head localization and incidence of skin cancers. *Scientific Reports*, *11*(1), 12391.

[22] Ali, A. R. H., Li, J., & Yang, G. (2020). Automating the ABCD rule for melanoma detection: a survey. *IEEE Access*, *8*, 83333-83346.

[23] M. D. Alahmadi, "Multiscale Attention U-Net for Skin Lesion Segmentation," in IEEE Access, vol. 10, pp. 59145-59154, 2022, doi: 10.1109/ACCESS.2022.3179390.

[24] Khouloud, S., Ahlem, M., Fadel, T. *et al.* W-net and inception residual network for skin lesion segmentation and classification. *Appl Intell* 52, 3976–3994 (2022). https://doi.org/10.1007/s10489-021-02652-4

[25] A. Karimi, K. Faez and S. Nazari, "DEU-Net: Dual-Encoder U-Net for Automated Skin Lesion Segmentation," in IEEE Access, vol. 11, pp. 134804-134821, 2023, doi: 10.1109/ACCESS.2023.3337528.

[26] S. Tasnim and M. F. Ahamed, "SEAA-UNet++: A Customized UNet++ Framework for Melanoma Segmentation in Dermoscopic Images with Test Time Augmentation," 2025 IEEE 7th International Conference on Sustainable Technologies For Industry 5.0 (STI), Dhaka, Bangladesh, 2025, pp. 1-6, doi: 10.1109/STI69347.2025.11367597.

[27] Hasan, M. K., Dahal, L., Samarakoon, P. N., Tushar, F. I., & Marti, R. (2020). DSNet: Automatic dermoscopic skin lesion segmentation. *Computers in biology and medicine*, *120*, 103738.

[28] M. Goyal, A. Oakley, P. Bansal, D. Dancey and M. H. Yap, "Skin Lesion Segmentation in Dermoscopic Images With Ensemble Deep Learning Methods," in IEEE Access, vol. 8, pp. 4171-4181, 2020, doi: 10.1109/ACCESS.2019.2960504.

[29] Sharen, H., Jawahar, M., Anbarasi, L. J., Ravi, V., Alghamdi, N. S., & Suliman, W. (2024). FDUM-Net: An enhanced FPN and U-Net architecture for skin lesion segmentation. *Biomedical Signal Processing and Control*, *91*, 106037. https://doi.org/10.1016/j.bspc.2024.106037

[30] Ali, M., Wu, T., Hu, H., Luo, Q., Xu, D., Zheng, W., Jin, N., Yang, C., & Yao, J. (2024). A review of the Segment Anything Model (SAM) for medical image analysis: Accomplishments and perspectives. *Computerized Medical Imaging and Graphics*, *119*, 102473. https://doi.org/10.1016/j.compmedimag.2024.102473

[31] Berrezueta, S., Baldeon-Calisto, M., Navarrete, D., Pérez-Pérez, N., Flores-Moyano, R., Riofrío, D., & Benítez, D. (2025, April). Foundation models for medical image segmentation: A literature review. In *2025 13th International Symposium on Digital Forensics and Security (ISDFS)* (pp. 1-7). IEEE.

[32] Wei, M., Chen, S., Wu, S., & Xu, D. (2025). Rep-MeDSAM: Towards Real-Time and Universal Medical Image Segmentation. In *Lecture notes in computer science* (pp. 57–69). https://doi.org/10.1007/978-3-031-81854-7_4

[33] Wei, X., Cao, J., Jin, Y., Lu, M., Wang, G., & Zhang, S. (2024). I-MeDSAM: Implicit Medical Image Segmentation with Segment Anything. *Lecture Notes in Computer Science*, 90–107. https://doi.org/10.1007/978-3-031-72684-2_6

[34] Huix, J. P., Ganeshan, A. R., Haslum, J. F., Söderberg, M., Matsoukas, C., & Smith, K. (2024). Are natural domain foundation models useful for medical image classification?. In *Proceedings of the IEEE/CVF winter conference on applications of computer vision* (pp. 7634-7643).

[35] Chen, C., Miao, J., Wu, D., Zhong, A., Yan, Z., Kim, S., Hu, J., Liu, Z., Sun, L., Li, X., Liu, T., Heng, P., & Li, Q. (2024). MA-SAM: Modality-agnostic SAM adaptation for 3D medical image segmentation. *Medical Image Analysis*, *98*, 103310. https://doi.org/10.1016/j.media.2024.103310

[36] Panhwar, A.O., Memon, S., Dhomeja, L.D., Memon, N. and Chandio, A.A., 2024. Deep Learning-Based Image Segmentation Techniques for Bone Fractures Using X-Ray Images: A Systematic Review. VFAST Transactions on Software Engineering, 12(4), pp. 99-116.

[37] Nigar, N., Umar, M., Shahzad, M. K., Islam, S., & Abalo, D. (2022). A deep learning approach based on explainable artificial intelligence for skin lesion classification. *IEEE Access*, *10*, 113715–113725. https://doi.org/10.1109/access.2022.3217217

[38] Aras, H. H., & Doğan, N. (2026). A comparative analysis for skin cancer detection by using explainable deep learning. *Neural Computing and Applications*, *38*(4). https://doi.org/10.1007/s00521-025-11809-y

[39] Alrabai, A., Echtioui, A., & Kallel, F. (2025). Explainable deep learning approaches for skin cancer diagnosis. *Network Modeling Analysis in Health Informatics and Bioinformatics*, *14*(1), 57.

[40] Mahmud, M. A. A., Afrin, S., Mridha, M. F., Alfarhood, S., Che, D., & Safran, M. (2025). Explainable deep learning approaches for high precision early melanoma detection using dermoscopic images. *Scientific Reports*, *15*(1), 24533.

[41] Khan, T., Haque, M. Z. U., Munir, G., & Usmani, I. A. (2026). Deep Learning-Based Skin Care Detection with Multi-method Explainability: Grad-CAM, Lime, and Occlusion Sensitivity. *IIUM Engineering Journal*, *27*(1), 160-174.

[42] Munjal, G., Bhardwaj, P., Bhargava, V., Singh, S., & Nagpal, N. (2024). SkinSage XAI: An explainable deep learning solution for skin lesion diagnosis. *Health Care Science*, *3*(6), 438-455.

[43] Codella, N., Rotemberg, V., Tschandl, P., Celebi, M. E., Dusza, S., Gutman, D., ... & Halpern, A. (2019). Skin lesion analysis toward melanoma detection 2018: A challenge hosted by the international skin imaging collaboration (isic). *arXiv preprint arXiv:1902.03368*.

[44] Mendonça, T., Ferreira, P. M., Marques, J. S., Marcal, A. R., & Rozeira, J. (2013, July). PH 2-A dermoscopic image database for research and benchmarking. In *2013 35th annual international conference of the IEEE engineering in medicine and biology society (EMBC)* (pp. 5437-5440). IEEE.

[45] Kingma, D. P., & Ba, J. (2014). Adam: A method for stochastic optimization. *arXiv preprint arXiv:1412.6980*.

[46] Lin, T. Y., Goyal, P., Girshick, R., He, K., & Dollár, P. (2017). Focal loss for dense object detection. In *Proceedings of the IEEE international conference on computer vision* (pp. 2980-2988).

[47] Zhang, J., Bargal, S. A., Lin, Z., Brandt, J., Shen, X., & Sclaroff, S. (2018). Top-down neural attention by excitation backprop. *International Journal of Computer Vision*, *126*(10), 1084-1102.

[48] Binzagr, F., & Hariri, M. (2026). Foundation-Model-Driven Skin Lesion Segmentation and Classification Using SAM-Adapters and Vision Transformers. *Diagnostics*, *16*(3), 468. https://doi.org/10.3390/diagnostics16030468

[49] Babu, E., Murali, S. AttenUNeT X with iterative feedback mechanisms for robust deep learning skin lesion segmentation. *Sci Rep* 15, 40690 (2025). https://doi.org/10.1038/s41598-025-23830-1

[50] A. Alrabai, A. Echtioui and F. Kallel, "Transformer and CNN-Based Deep Learning Models for Skin Lesion Segmentation," *2025 IEEE International Conference on Advanced Systems and*

*Emergent Technologies (IC_ASET)*, Mammamet-Yasmine, Tunisia, 2025, pp. 1-6, doi: 10.1109/IC_ASET65966.2025.11232106.

[51] M. Mushgil, H. ., Saad Al-Mukhtar, F. ., Qahtan Ahmed, . E. ., & Saeed Abduljabbar, K. . (2025). Deep Image Segmentation Using Explainable Attention Mechanisms: Applications in Biomedical Imaging. *Al-Nahrain Journal of Science*, *28*(4), 241-258. https://anjs.edu.iq/index.php/anjs/article/view/3215

[52] W. Chen, Q. Mu and J. Qi, "TrUNet: Dual-Branch Network by Fusing CNN and Transformer for Skin Lesion Segmentation," in *IEEE Access*, vol. 12, pp. 144174-144185, 2024, doi: 10.1109/ACCESS.2024.3463713.

[53] Liu, X., Gao, P., Yu, T., Wang, F., & Yuan, R. (2024). CSWin-UNet: Transformer UNet with cross-shaped windows for medical image segmentation. *Information Fusion*, *113*, 102634. https://doi.org/10.1016/j.inffus.2024.102634

[54] Heidari, M., Kazerouni, A., Soltany, M., Azad, R., Aghdam, E. K., Cohen-Adad, J., & Merhof, D. (2023). Hiformer: Hierarchical multi-scale representations using transformers for medical image segmentation. In *Proceedings of the IEEE/CVF winter conference on applications of computer vision* (pp. 6202-6212).

[55] Hu, M., Li, Y., & Yang, X. (2023). Skinsam: Empowering skin cancer segmentation with segment anything model. *arXiv preprint arXiv:2304.13973*.

[56] Li, Z., Zhang, H., Li, Z., & Ren, Z. (2022). Residual-Attention UNet++: A Nested Residual-Attention U-Net for Medical Image Segmentation. *Applied Sciences*, *12*(14), 7149. https://doi.org/10.3390/app12147149

[57] Abraham, N., & Khan, N. M. (2019, April). A novel focal tversky loss function with improved attention u-net for lesion segmentation. In *2019 IEEE 16th international symposium on biomedical imaging (ISBI 2019)* (pp. 683-687). IEEE.

[58] Tong, X., Wei, J., Sun, B., Su, S., Zuo, Z., & Wu, P. (2021). ASCU-Net: Attention Gate, Spatial and Channel Attention U-Net for Skin Lesion Segmentation. *Diagnostics*, *11*(3), 501. https://doi.org/10.3390/diagnostics11030501

[59] Luo, W. (2026). Skin Lesion Segmentation via Improved U-Net with Spatial Group-wise Enhancement and Multi-scale Parallel Feature Fusion. *Journal of Computing and Electronic Information Management*, *20*(2), 81-84. https://doi.org/10.54097/dva4jf40

[60] Kaur, R., GholamHosseini, H., Sinha, R. *et al.* Automatic lesion segmentation using atrous convolutional deep neural networks in dermoscopic skin cancer images. *BMC Med Imaging* **22**, 103 (2022). https://doi.org/10.1186/s12880-022-00829-y

**[61]** Wang, M., Liang, Y., Tang, Y., Feng, Y., Zhang, T., & Lv, C. (2024, November). Global-local medical sam adaptor based on full adaption. In *2024 2nd International Conference on Computer, Vision and Intelligent Technology (ICCVIT)* (pp. 1-6). IEEE.